\documentclass[letterpaper, 10 pt, journal, twoside]{ieeetran}  

\IEEEoverridecommandlockouts                              

\usepackage{amsmath} 
\usepackage{amssymb}  

\usepackage{verbatim}
\usepackage{cite}

\usepackage{mathtools}
\usepackage[ruled,linesnumbered]{algorithm2e}

\usepackage{amssymb}

\usepackage{rotating}
\usepackage{graphicx}
\usepackage{caption}
\usepackage{subcaption}
\usepackage{comment}
\usepackage[colorinlistoftodos]{todonotes}

\usepackage{ulem}

\usepackage{fancybox,color}

\definecolor{grey}{rgb}{0.5,0.5,0.5}

\usepackage{hyperref} 

\usepackage[utf8]{inputenc}
\usepackage[english]{babel}

\usepackage{booktabs}
\usepackage{multirow}
\usepackage{tikz}

\title{The Multi Vehicle Stereo Event Camera Dataset: \\An Event Camera Dataset for 3D Perception}

\author{Alex Zihao Zhu$^{1}$, Dinesh Thakur$^{1}$, Tolga \"Ozaslan$^{1}$, Bernd Pfrommer$^{1}$, Vijay Kumar$^{1}$ and Kostas Daniilidis$^{1}$
\thanks{Manuscript received: September, 11, 2017; Revised December, 8, 2017; Accepted January, 13, 2018.}
\thanks{This paper was recommended for publication by Editor Cyrill Stachniss upon evaluation of the Associate Editor and Reviewers' comments. This work was supported by NSF-DGE-0966142 (IGERT), NSF-IIP-1439681 (I/UCRC), NSF-IIS-1426840, ARL MAST-CTA W911NF-08-2-0004, ARL RCTA W911NF-10-2-0016, ONR N00014-17-1-2093, an ONR STTR (Robotics Research), NSERC Discovery, and the DARPA FLA program.}
\thanks{$^{1}$A. Z. Zhu, D. Thakur, T. \"Ozaslan, B. Pfrommer, V. Kumar and K. Daniilidis are with the GRASP Laboratory,
        University of Pennsylvania, Pennsylvania, PA 19104
        {\tt\small (alexzhu, tdinesh, ozaslan, pfrommer, kumar, kostas)@seas.upenn.edu}}%
\thanks{Digital Object Identifier (DOI): see top of this page.}
}

\begin{document}
\maketitle

\begin{abstract}
Event based cameras are a new passive sensing modality with a number of benefits over traditional cameras, including extremely low latency, asynchronous data acquisition, high dynamic range and very low power consumption. There has been a lot of recent interest and development in applying algorithms to use the events to perform a variety of 3D perception tasks, such as feature tracking, visual odometry, and stereo depth estimation. However, there currently lacks the wealth of  labeled data that exists for traditional cameras to be used for both testing and development. In this paper, we present a large dataset with a synchronized stereo pair event based camera system, carried on a handheld rig, flown by a hexacopter, driven on top of a car and mounted on a motorcycle, in a variety of different illumination levels and environments. From each camera, we provide the event stream, grayscale images and IMU readings. In addition, we utilize a combination of IMU, a rigidly mounted lidar system, indoor and outdoor motion capture and GPS to provide accurate pose and depth images for each camera at up to 100Hz. For comparison, we also provide synchronized grayscale images and IMU readings from a frame based stereo camera system.
\end{abstract}
\begin{IEEEkeywords}
SLAM, Visual-Based Navigation, Event-Based Cameras
\end{IEEEkeywords}

\section{INTRODUCTION}
\IEEEPARstart{E}{VENT} based cameras sense the world by detecting changes in the log intensity of an image. 
By registering these changes with accuracy on the order of tens of microseconds and asynchronous, almost instant, feedback, they allow for extremely low latency responses compared to traditional cameras which typically have latencies on the order of tens of milliseconds. In addition, by tracking changes in log intensity, the cameras have very high dynamic range ($>$130dB vs about 60dB with traditional cameras), which make them very useful for scenes with dramatic changes in lighting, such as indoor-outdoor transitions, as well as scenes with a strong light source, such as the sun. 

However, most modern robotics algorithms have been designed for synchronous sensors, where measurements arrive at fixed time intervals. In addition, the generated events do not carry any intensity information on their own. As a result, new algorithms must be developed to fully take advantage of the benefits provided by this sensor. Unfortunately, due to the differences in measurements, we cannot directly take advantage of the enormous amounts of labeled data captured with traditional cameras. Such data has shown to be extremely important for providing realistic and consistent evaluations of new methods, training machine learning systems, and providing opportunities for new development for researchers who do not have access to these sensors.

\begin{figure}[t]
\centering
  \includegraphics[width=0.8\linewidth]{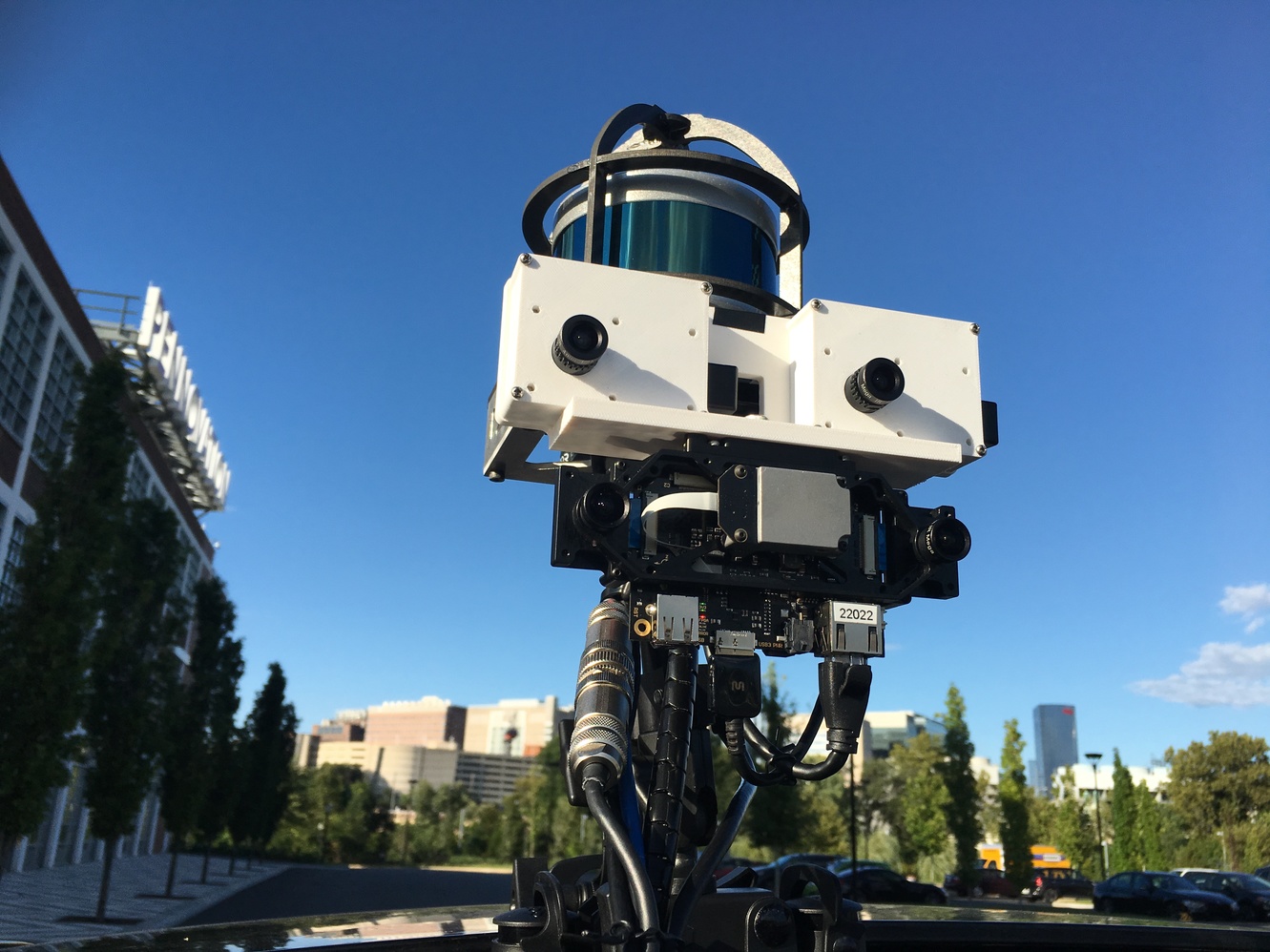}
  \caption{Full sensor rig, with stereo DAVIS cameras, VI Sensor and Velodyne lidar.}
  \vspace{-0.5cm}
  \label{fig:full_rig}
\end{figure}

In this work, we aim to provide a number of different sequences that will facilitate the research and development of novel solutions to a number of different problems. One main contribution is the first dataset with a synchronized stereo event camera system. A calibrated stereo system is useful for depth estimation with metric scale, which can contribute to problems such as pose estimation, mapping, obstacle avoidance and 3D reconstruction. There have been a few works in stereo depth estimation with event based cameras, but, due to the lack of accurate ground truth depth, the evaluations have been limited to small, disparate sequences, consisting of a few objects in front of the camera. In comparison, this dataset provides event streams from two synchronized and calibrated Dynamic Vision and Active Pixel Sensors (DAVIS-m346b), with long indoor and outdoor sequences in a variety of illuminations and speeds, along with accurate depth images and pose at up to 100Hz, generated from a lidar system rigidly mounted on top of the cameras, as in Fig~\ref{fig:full_rig}, along with motion capture and GPS. We hope that this dataset can help provide a common basis for event based algorithm evaluation in a number of applications.

The full dataset can be found online at \url{https://daniilidis-group.github.io/mvsec}.
\newpage
The main contributions from this paper can be summarized as:
\begin{itemize}
\item
The first dataset with synchronized stereo event cameras, with accurate ground truth depth and pose.
\item
Event data from a handheld rig, a flying hexacopter, a car, and a motorcycle, in conjunction with calibrated sensor data from a 3D lidar, IMUs and frame based images, from a variety of different speeds, illumination levels and environments.
\end{itemize}

\section{RELATED WORK}
\subsection{Related Datasets}

At present, there are a number of existing datasets that provide events from monocular event based cameras in conjunction with a variety of other sensing modalities and ground truth measurements that are suitable for testing a number of different 3D perception tasks. 

Weikersdorfer et al. \cite{weikersdorfer2014event} combine the earlier eDVS sensor with 128x128 resolution, with a Primesense RGBD sensor, and provide a dataset of indoor sequences with ground truth pose from a motion capture system, and depth from the RGBD sensor.

Rueckauer et al. \cite{rueckauer2016evaluation} provide data from a DAVIS 240C camera undergoing pure rotational motion, as well as ground truth optical flow based on the angular velocities reported from the gyroscope, although this is subject to noise in the reported velocities.

Barranco et al. \cite{barranco2016dataset} present a dataset with a DAVIS 240B camera mounted on top of a pan tilt unit, attached to a mobile base, along with a Microsoft Kinect sensor. The dataset provides sequences of the base moving with 5dof in a indoor environment, along with ground truth depth, and optical flow and pose from the wheel encoders on the base and the angles from the pan tilt unit. While the depth from the Kinect is accurate, the optical flow and pose are subject to drift from the position estimates of the base's wheel encoders. 

Mueggler et al. \cite{eventcameradataset} provide a number of handheld sequences intended for pose estimation in a variety of indoor and outdoor environments, generated from a DAVIS 240C. A number of the indoor scenes have provided pose ground truth, captured from a motion capture system. However, there are no outdoor sequences, or other sequences with a significant displacement, with ground truth information.

Binas et al. \cite{binas2017ddd17} provide a large dataset of a DAVIS 346B mounted behind the windshield of a car, with 12 hours of driving, intended for end to end learning of various driving related tasks. The authors provide a number of auxiliary measurements from the vehicle, such as steering angle, accelerator pedal position, vehicle speed etc., as well as longitude and latitude from a GPS unit. However, no 6dof pose is provided, as only 2D translation can be inferred from the GPS output as provided.

These datasets provide valuable data for development and evaluation of event based methods. However, they have, to date, only monocular sequences, with ground truth 6dof pose limited to small indoor environments, with few sequences with ground truth depth. In contrast, this work provides stereo sequences with ground truth pose and depth images in a variety of indoor and outdoor settings.
\subsection{Event Based 3D Perception}
Early works in \cite{kogler2010address}, \cite{kogler2011event} present stereo depth estimation results with a number of spatial and temporal costs. Later works in 
\cite{piatkowska2014cooperative}, \cite{Firouzi2016} and \cite{piatkowska2017improved} have adapted cooperative methods for stereo depth to event based cameras, due to their applicability to asynchronous, point based measurements. Similarly, \cite{rogister2012asynchronous} and \cite{carneiro2013event} apply a set of temporal, epipolar, ordering and polarity constraints to determine matches, while \cite{camunas2014use} compare this with matching based on the output of a bank of orientation filters. The authors in \cite{benosman2011asynchronous} show a new method to determine the epipolar line, applied to stereo matching. In \cite{zou2016context}, the authors propose a novel context descriptor to perform matching, and the authors in \cite{schraml2015event} use a stereo event camera undergoing pure rotation to perform depth estimation and panoramic stitching. 

There are also a number of works on event based visual odometry and SLAM problems. The authors in \cite{zhu2017event} and \cite{tedaldi2016feature} proposed novel methods to perform feature tracking in the event space, which they extended in \cite{kueng2016low} and \cite{zhuevent} to perform visual and visual inertial odometry, respectively. In \cite{weikersdorfer2014event}, the authors combine an event based camera with a depth sensor, to perform visual odometry and SLAM. The authors in \cite{gallego2017accurate} use events to estimate angular velocity of a camera, while \cite{kim2016real} and \cite{rebecq2017evo} perform visual odometry by building an up to a scale map. In addition, \cite{rebecq2017real} and \cite{mueggler2017continuous} also fuse events with measurements from an IMU to perform visual inertial odometry.

While the more recent works evaluate based on public datasets such as \cite{eventcameradataset}, the majority are evaluated on small datasets generated solely for the paper, making comparisons of performance difficult. This is particularly the case for stereo event based cameras. In this work, we try to generate more extensive ground truth, for more meaningful evaluations of new algorithms that can provide a basis for comparisons between methods.
\begin{figure*}[t]
\centering
  \begin{subfigure}[b]{0.28\textwidth}
  \centering
    \includegraphics[width=\linewidth]{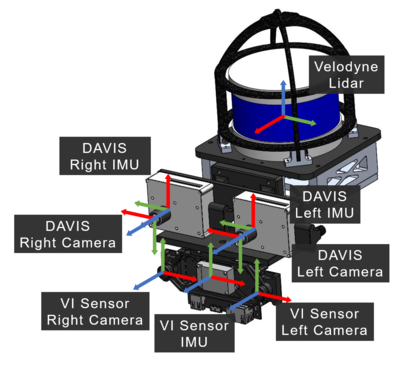}
    \caption{}
    \label{fig:rig_frames}
  \end{subfigure}
  ~
  \begin{subfigure}[b]{0.24\textwidth}
  \centering
  \includegraphics[width=\linewidth]{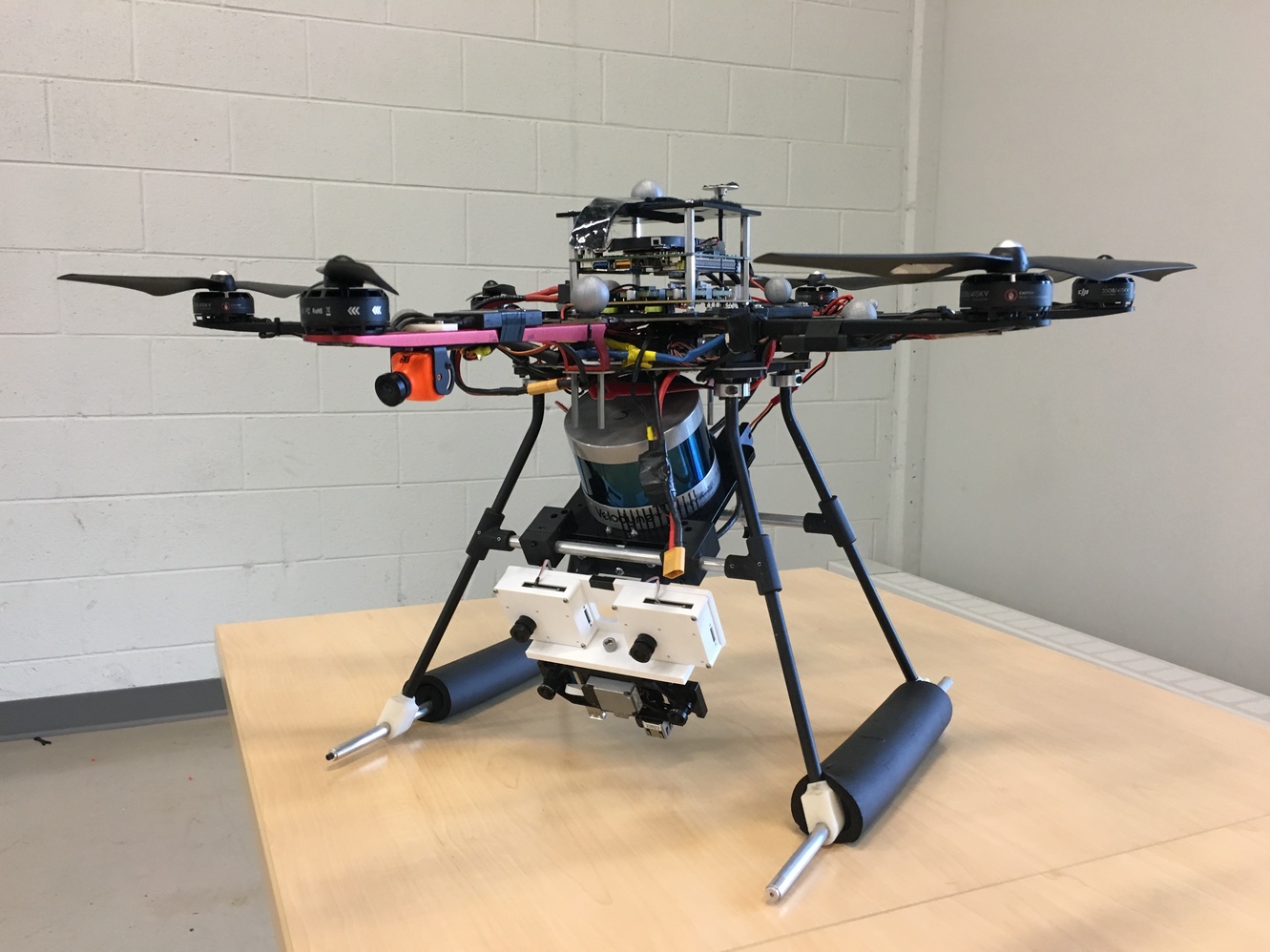}
  \caption{}
  \label{fig:quad}
  \end{subfigure}
  ~
  \begin{subfigure}[b]{0.18\textwidth}
  \centering
  \begin{turn}{90}
  \includegraphics[width=\linewidth]{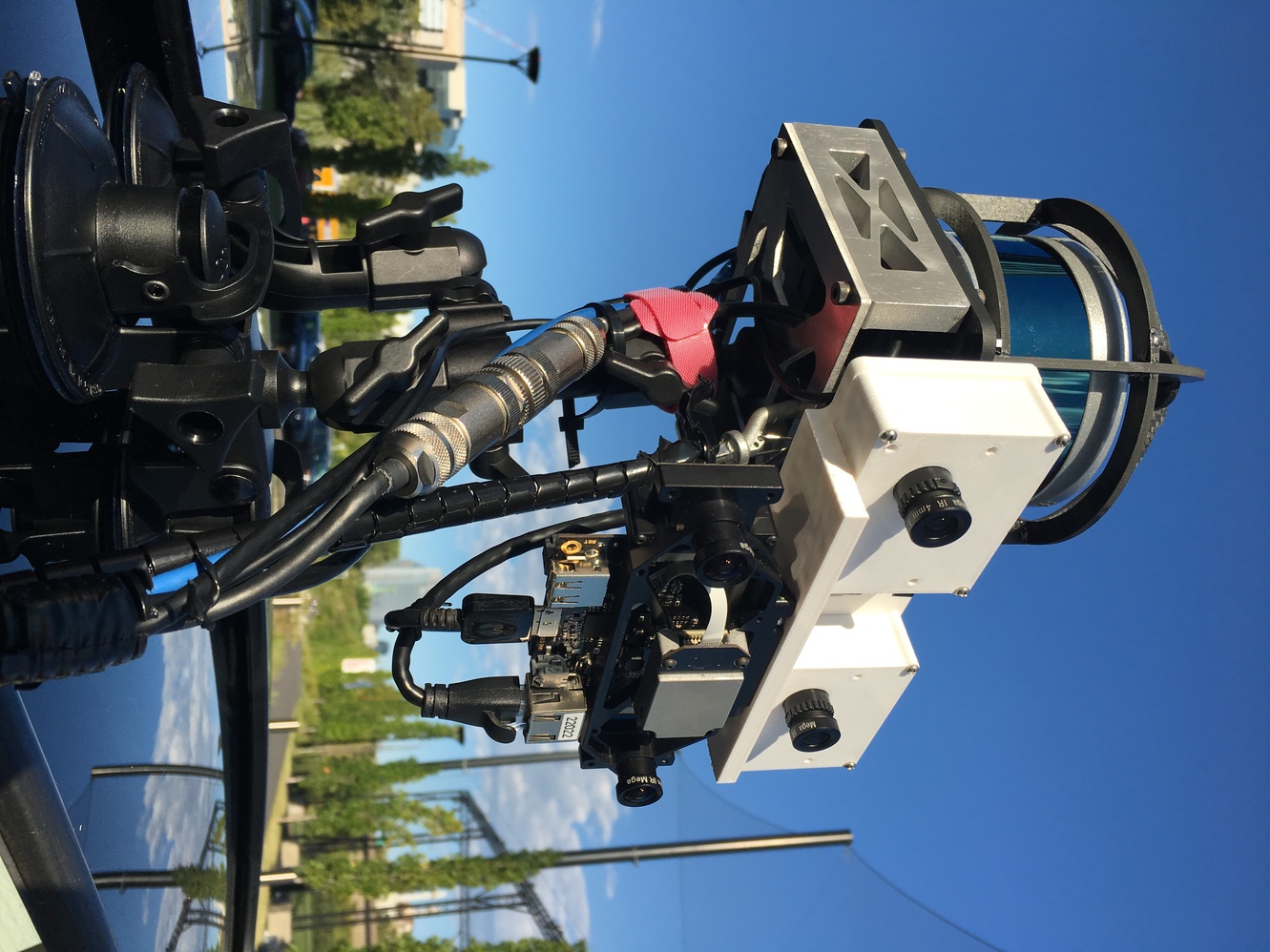}
  \end{turn}
  \caption{}
  \label{fig:car_mount}
  \end{subfigure}
  ~
  \begin{subfigure}[b]{0.24\textwidth}
  \centering
  \includegraphics[width=\linewidth]{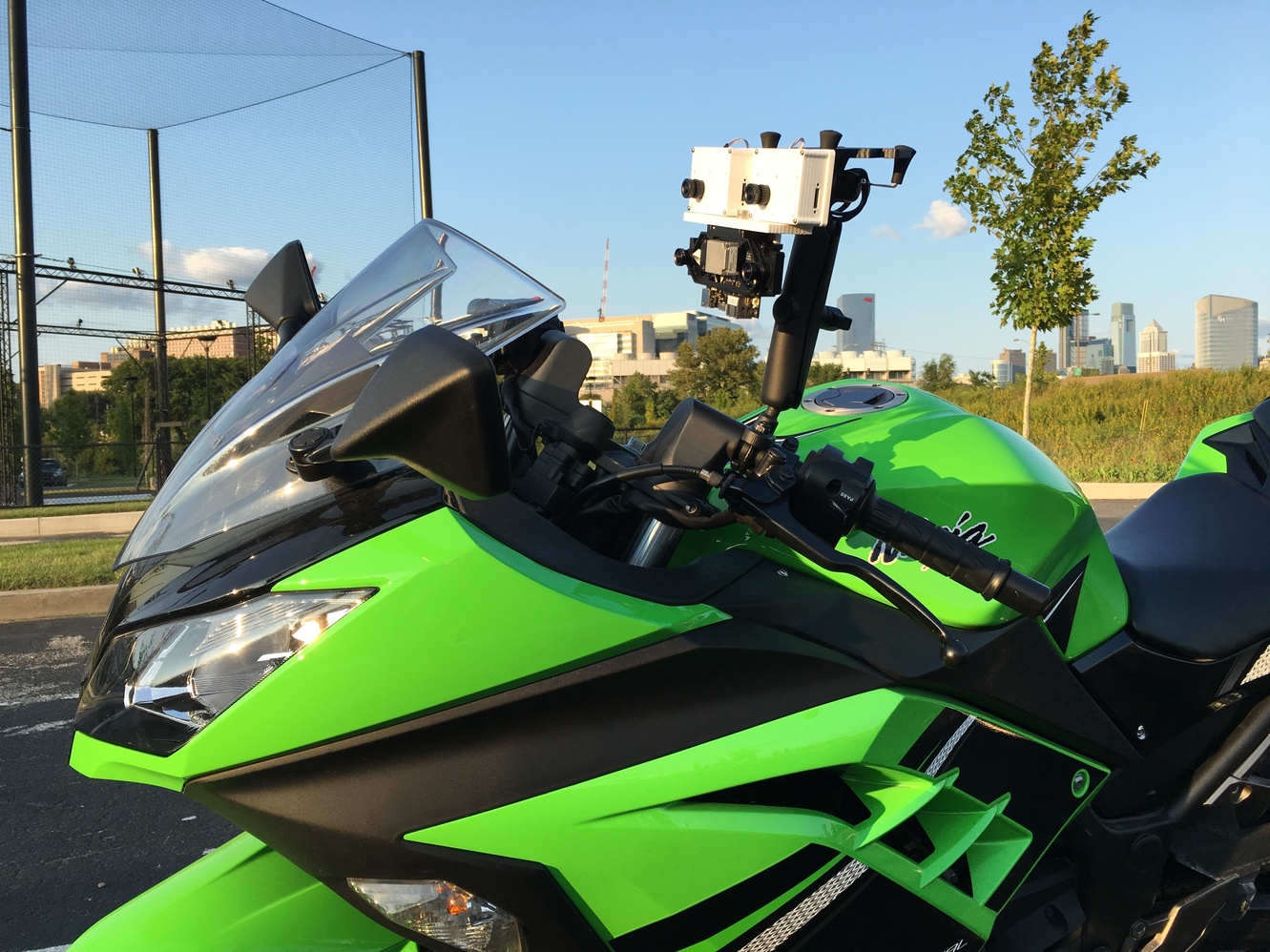}
  \caption{}
  \label{fig:motorbike_mount}
  \end{subfigure}
  \caption{Left to right: (a): CAD model of the sensor rig. All sensor axes are labeled and colored R:X, G:Y, B:Z, with only combinations of approximately 90 degree rotations between each pair of axes. (b): Sensor package mounted on hexacopter. (c): Sensor package mounted using a glass suction tripod mount on the sunroof of a car. (d): DAVIS cameras and VI Sensor mounted on motorcycle. Note that the VI-Sensor is mounted upside down in all configurations. Best viewed in color.}
  \label{fig:vehicles}
\end{figure*}

\section{DATASET} \label{sec:DataSet}
For each sequence in this dataset, we provide the following measurements in ROS bag\footnote{\url{http://wiki.ros.org/Bags}} format:
\begin{itemize}
\item Events, APS grayscale images and IMU measurements from the left and right DAVIS cameras.
\item Images and IMU measurements from the VI Sensor.
\item Pointclouds from the Velodyne VLP-16 lidar.\footnote{\url{http://velodynelidar.com/vlp-16-lite.html}}
\item Ground truth reference poses for the left DAVIS camera.
\item Ground truth reference depth images for both left and right DAVIS cameras.
\end{itemize}
\subsection{Sensors}
\begin{table}[h]
\centering
\begin{tabular}{@{}cl@{}}
\toprule
Sensor                              & Characteristics                                                           \\ \midrule
\multirow{3}{*}{DAVIS m346B} & 346x260 pixel APS+DVS                                                                   \\
                                    & FOV: 67$^{\circ}$ vert., 83$^{\circ}$ horiz.                              \\
                                    & IMU: MPU 6150
                                    \\ \midrule
\multirow{5}{*}{VI-Sensor}          & Skybotix integrated VI-sensor                                             \\
                                    & stereocamera: 2 Aptina MT9V034                                            \\
                                    & gray 2x752x480 @ 20fps, global shutter                        \\
                                    & FOV: 57$^{\circ}$ vert., 2 x 80$^{\circ}$ horiz.                          \\
                                    & IMU: ADIS16488 @200Hz                                                     \\ \midrule
\multirow{5}{*}{Velodyne Puck LITE} & VLP-16 PUCK LITE                                                          \\
                                    & 360$^{\circ}$ Horizontal FOV, 30$^{\circ}$ Vertical FOV \\
                                    & 16 channel                                                                \\
                                    & 20Hz                                                                      \\
                                    & 100m Range                                                                \\ \midrule
GPS                                 & UBLOX NEO-M8N                                                     		\\
                                    & 72-channel u-blox M8 engine                                               \\
                                    & Position accuracy 2.0 m CEP                                                \\ \midrule
                                    &                                                                          
\end{tabular}
\caption{Sensors and characteristics.}
\label{tab:sens_char}
\vspace{-10pt}
\end{table}

\begin{table*}[t]
\vspace{5pt}
\centering
\begin{tabular}{llcccccll}
\toprule
Vehicle & Sequence         	& T(s) & D(m) & $\|v\|_{\text{max}}$(m/s) & $\|\omega\|_{\text{max}}$($^\circ$/s) & MER(events/s) & Pose GT & Depth Available\\ 
\midrule
Hexacopter & Indoor 1$^*$ & 70 & 26.7 & 1.4 & 28.3 & 185488 & Vicon & Yes\\ 
& Indoor 2$^*$ & 84 & 36.8 & 1.5 & 29.8 & 273567 & Vicon & Yes\\
& Indoor 3$^*$ & 94 & 52.3 & 1.7& 31.0 & 243953 & Vicon & Yes\\
& Indoor 4$^*$ & 20 & 9.8 & 2.0 & 64.3 & 361579 & Vicon & Yes\\
& Outdoor 1 & 54 & 33.2 & 1.5 & 129.8 & 261589 & Qualisys & Yes\\
& Outdoor 2	& 41 & 29.9 & 1.5 & 109.4 & 256539 & Qualisys & Yes\\\midrule
Handheld & Indoor-outdoor & 144 & 80.4 & 1.6 & 93.6 & 468675 & LOAM & Yes\\
& Indoor corridor & 249 & 105.2 & 2.7 & 37.4 & 590620 & LOAM & Yes\\\midrule
Car & Day 1$^\dagger$ & 262 & 1207 & 7.6 & 30.6 & 386178 & Cart., GPS & Yes\\
& Day 2$^\dagger$ & 653 & 3467 & 12.0 & 35.5 & 649081 & Cart., GPS & Yes\\
& Evening 1 & 262 & 1217 & 10.4 & 20.6 & 334614 & Cart., GPS & Yes\\
& Evening 2 & 374 & 2109 & 11.2 & 33.6 & 404105 & Cart., GPS & Yes\\
& Evening 3	& 276 & 1613 & 10.0 & 25.1 & 371498 & Cart., GPS & Yes\\\midrule
Motorcycle & Highway 1 & 1500 & 18293 & 38.4 & 203.4 & 511024 & GPS & No\\\midrule
\end{tabular}
\caption{Sequences for each vehicle. T: Total time, D: Total distance traveled, $\|v\|_{\text{max}}$: Maximum linear velocity, $\|\omega\|_{\text{max}}$: Maximum angular velocity, MER: Mean event rate.\\
$^*$No VI-Sensor data is available for these sequences.\\
$^\dagger$A hardware failure caused the right DAVIS grayscale images to fail for these sequences.}
\label{tab:sequences}
\end{table*}

A list of sensors and their characteristics can be found in Table~\ref{tab:sens_char}. In addition, Fig~\ref{fig:rig_frames} shows the CAD drawing of the sensor rig, with all sensor axes labeled, and Fig~\ref{fig:vehicles} shows how the sensors are mounted on each vehicle. The extrinsics between all sensors are estimated through calibration, as explained in Sec~\ref{sec:calibration}.

For event generation, two experimental mDAVIS-346B cameras are mounted in a horizontal stereo setup. The cameras are similar to \cite{brandli2014240}, but have a higher, 346x260 pixel, resolution, up to 50fps APS (frame based images) output, and higher dynamic range. The baseline of the stereo rig is 10cm, and the cameras are timestamp synchronized by using the trigger signal generated from the left camera (master) to deliver sync pulses to the right (slave) through an external wire. Both cameras have 4mm lenses with approximately 87 degrees horizontal field of view, with an additional IR cut filter placed on each one to suppress the IR flashes from the motion capture systems. The APS exposures are manually set (no auto exposure) depending on lighting conditions, but are always the same between the cameras. While the timestamps of the grayscale DAVIS images are synced, there is unfortunately no way to synchronize the image acquisition itself. Therefore, there may be up to 10ms of offset between the images.

To provide ground truth reference poses and depths (Sec \ref{sec:groundtruth}), we have rigidly mounted a Velodyne Puck LITE above the stereo DAVIS cameras. The Velodyne lidar system provides highly accurate depth of a large number of points around the sensor. The lidar is mounted such that there is full overlap between the smaller vertical field of view of the lidar and that of the stereo DAVIS rig. 

In the outdoor scenes, we have also mounted a GPS device for a second ground truth reference for latitude and longitude. Typically, the GPS is placed away from the sensor rig to avoid interference from the USB 3.0 data cables.

In addition, we have mounted a VI Sensor \cite{visensor}, originally developed by Skybotix for comparison with frame based methods. The sensor has a stereo pair with IMU, all synchronized. Unfortunately, the only mounting option was to mount the cameras upside down, but we provide the transform between them and the DAVIS cameras. 
\begin{figure*}[h]
    \centering
    \begin{subfigure}[t]{0.26\textwidth}
        \centering
        \includegraphics[trim=0 0 0 -1cm, width=\linewidth]{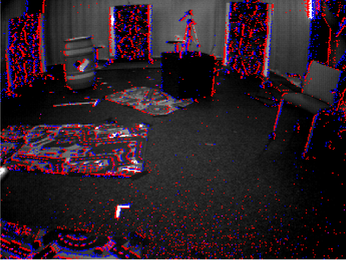}
        \caption{Hexacopter Indoor Flight with Vicon Motion Capture.}
    \end{subfigure}%
    ~
    \begin{subfigure}[t]{0.26\textwidth}
        \centering
        \includegraphics[trim=0 0 0 -1cm, width=\linewidth]{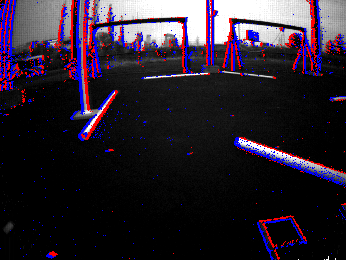}
        \caption{Hexacopter Outdoor Flight with Qualisys Motion Capture.}
    \end{subfigure}%
    ~
    \begin{subfigure}[t]{0.26\textwidth}
        \centering
        \includegraphics[width=\linewidth]{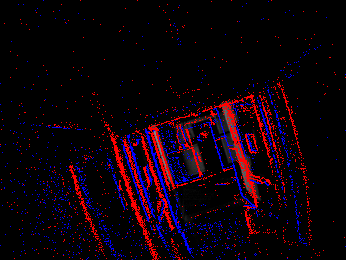}
        \caption{Handheld with Difficult Lighting Conditions.}
    \end{subfigure}%
    
    \begin{subfigure}[t]{0.26\textwidth}
        \centering
        \includegraphics[trim=0 0 0 -1cm, width=\linewidth]{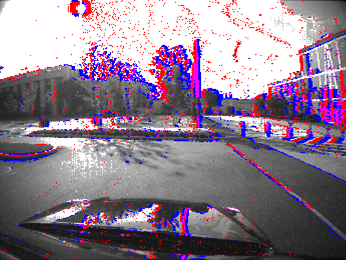}
        \caption{Car Day 1.}
    \end{subfigure}%
 	~
    \begin{subfigure}[t]{0.26\textwidth}
        \centering
        \includegraphics[trim=0 0 0 -1cm, width=\linewidth]{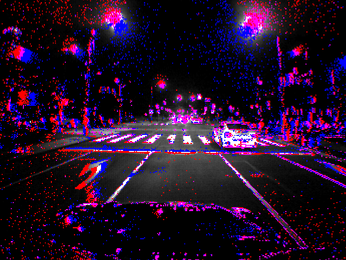}
        \caption{Outdoor Car Evening.}
    \end{subfigure}
    ~
    \begin{subfigure}[t]{0.26\textwidth}
        \centering
        \includegraphics[trim=0 0 0 -1cm, width=\linewidth]{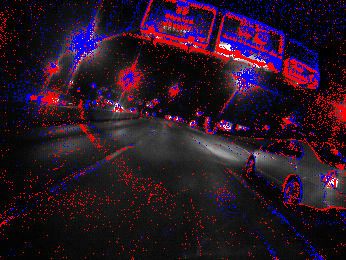}
        \caption{Motorcycle Highway 1.}
    \end{subfigure}
    \caption{Sample images with overlaid events (blue and red) from indoor and outdoor sequences, during day and evening. Best viewed in color.}
    \label{fig:example_images}
\end{figure*}
\begin{figure*}[h!]
\centering
  \includegraphics[width=0.33\linewidth]{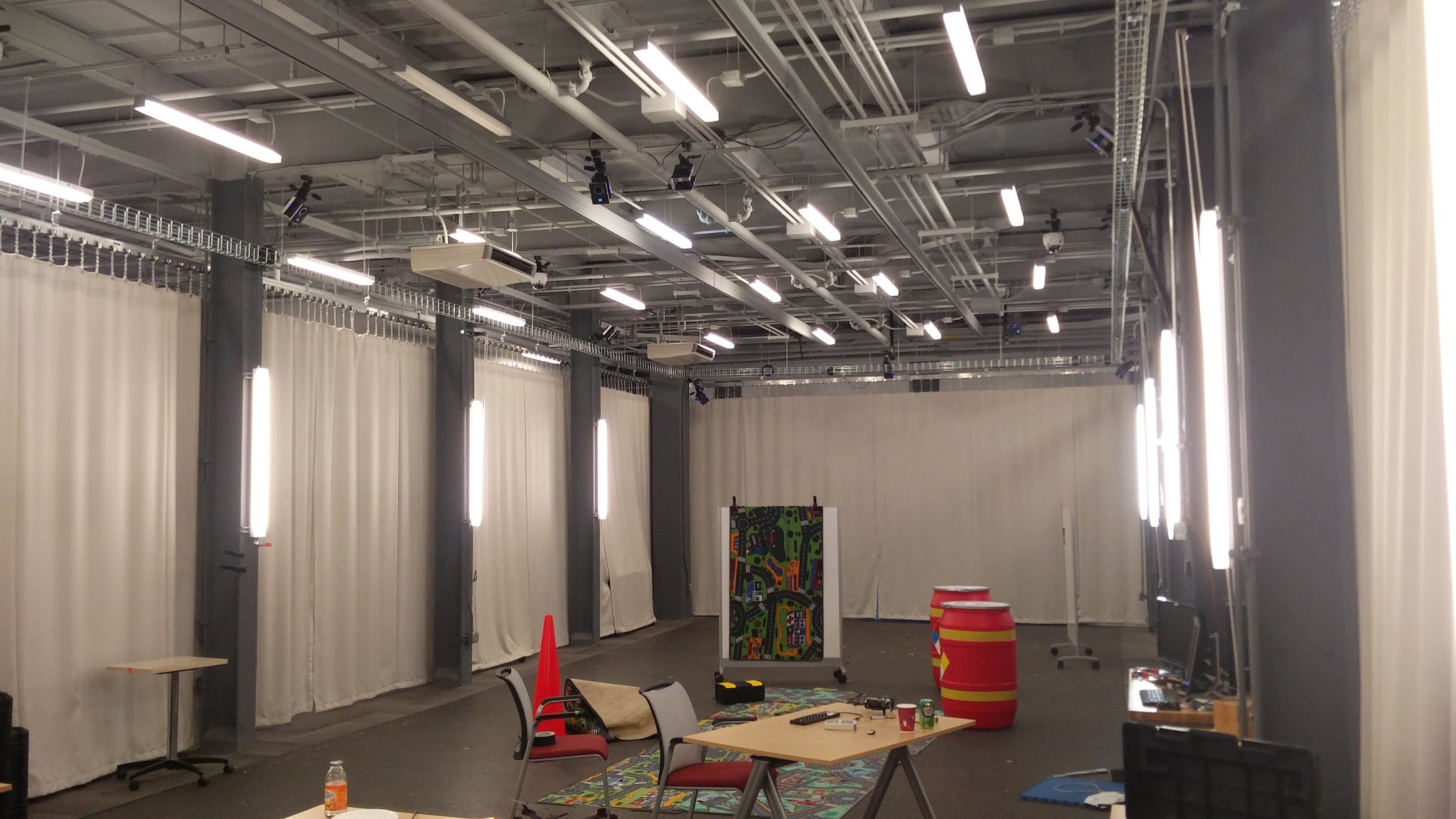}
  \includegraphics[width=0.25\linewidth]{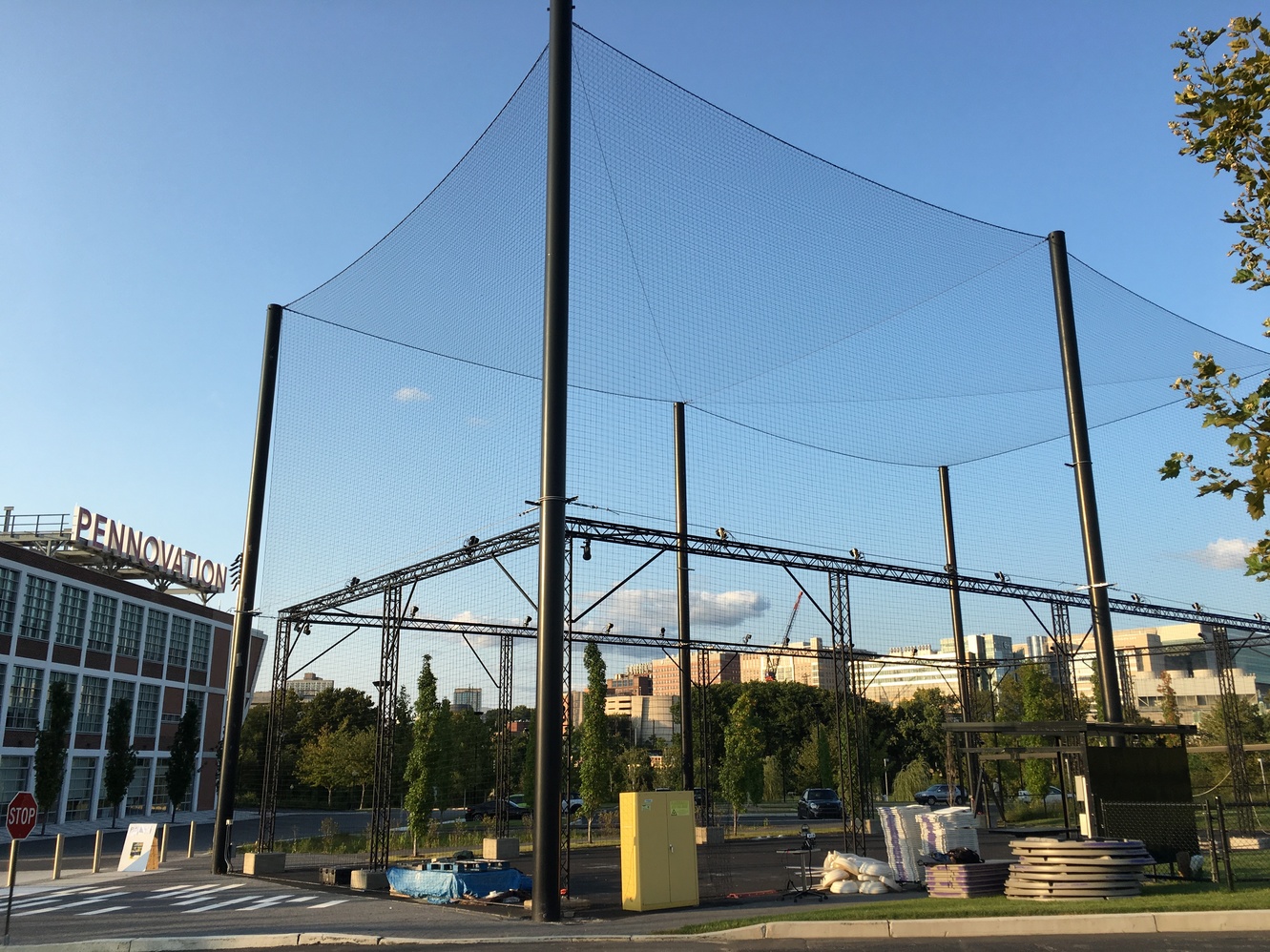}
  \caption{Motion capture arenas. Left: Indoor Vicon arena, right: Outdoor Qualisys arena.}
  \label{fig:mocap}
  \vspace{-10pt}
\end{figure*}
\subsection{Sequences}
The full list of sequences with summary statistics can be found in Table~\ref{tab:sequences}, and sample APS images with overlaid events can be found in Fig~\ref{fig:example_images}.

\subsubsection{Hexacopter with Motion Capture}
The sensor setup was mounted below the compute stack of the hexacopter, with a 25 degree downwards pitch, as in Fig~\ref{fig:quad}. Two motion capture systems are used to generate sequences for this dataset, one indoors and one outdoors (Fig~\ref{fig:mocap}). The $\text{26.8m} \times \text{6.7m} \times \text{4.6m}$ indoor area is instrumented with 20 Vicon Vantage VP-16 cameras. The outdoor netted area of $\text{30.5m} \times \text{15.3m} \times \text{15.3m}$ is instrumented with an all-weather motion capture system comprised of 34 high resolution Qualisys Oqus 700 cameras. Both systems provide millimeter accuracy pose at 100Hz by emitting infrared strobes and tracking IR reflecting markers placed on the hexacopter. We provide sequences in each area, with flights of different length and speed.

\subsubsection{Handheld}
In order to test performance in high dynamic range scenarios, the full sensor rig is carried in a loop through both outdoor and indoor environments, as well as indoor environments with and without external lighting. Ground truth pose and depth is provided by lidar SLAM.

\subsubsection{Outdoor Driving}
For slow to medium speed sequences, the sensor rig is mounted on the sun roof of a sedan as in Fig~\ref{fig:car_mount}, and driven around several West Philadelphia neighborhoods at speeds up to 12 m/s. Sequences are provided in both day and evening situations, including sequences with the sun directly in the cameras' field of view. Ground truth is provided as depth images from a lidar map, as well as pose from loop closed lidar odometry and GPS.

For high speed sequences, the DAVIS stereo rig and VI Sensor are mounted on the handlebar of a motorcycle (Fig~\ref{fig:motorbike_mount}), along with the GPS device. The sequences involve driving at up to 38m/s. Longitude and latitude, as well as relative velocity, are provided from the GPS.
\section{GROUND TRUTH GENERATION} \label{sec:groundtruth}
To provide ground truth poses, motion capture poses are used when available. Otherwise, if lidar is available, Cartographer \cite{cartographer} is used for the driving sequences to fuse the lidar sweeps and IMU data into a loop-closed 2D pose of the lidar, which is transformed into the left DAVIS frame using the calibration in Sec \ref{sec:lidartocamcalib}. For outdoor scenes, we also provide raw GPS readings.

For each sequence with lidar measurements, we run the Lidar Odometry and Mapping (LOAM) algorithm \cite{loam} to generate dense 3D local maps, which are projected into each DAVIS camera to generate dense depth images at 20Hz, and to provide 3D pose for the handheld sequences. 

Two separate lidar odometry algorithms are used as we noted that LOAM produces better, more well aligned, local maps, while Cartographer's loop closure results in more accurate global poses with less drift for longer trajectories. While Cartographer only estimates a 2D pose, we believe that this is a valid assumption as the roads driven have, for the most part, a single consistent grade.

\subsection{Ground Truth Pose}
For the sequences in the indoor and outdoor motion capture arenas, the pose of the body frame of the sensor rig $^{\text{world}}\mathbf{H}_{\text{body}(t)}$ at each time $t$ is measured at 100Hz with millimeter level accuracy.

For outdoor sequences we rely on Cartographer to perform loop closure and fuse lidar sweeps and IMU data into a single loop-closed 2D pose of the body (lidar in this case) with minimal drift. 

\begin{figure*}[h!]
\centering
  \begin{subfigure}[b]{0.45\textwidth}
   \centering 
  \includegraphics[width=0.73\linewidth]{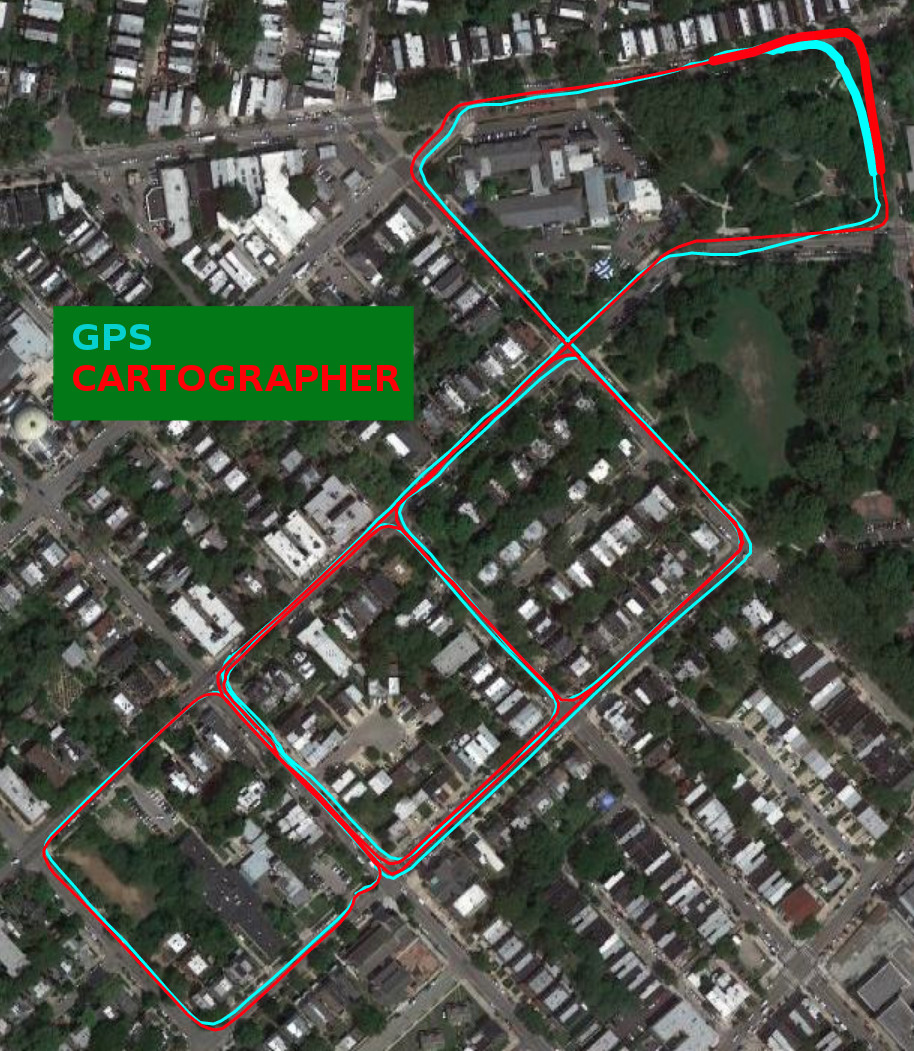}
  \end{subfigure}
  ~
  \begin{subfigure}[b]{0.45\textwidth}
  \centering
  \includegraphics[width=0.88\linewidth]{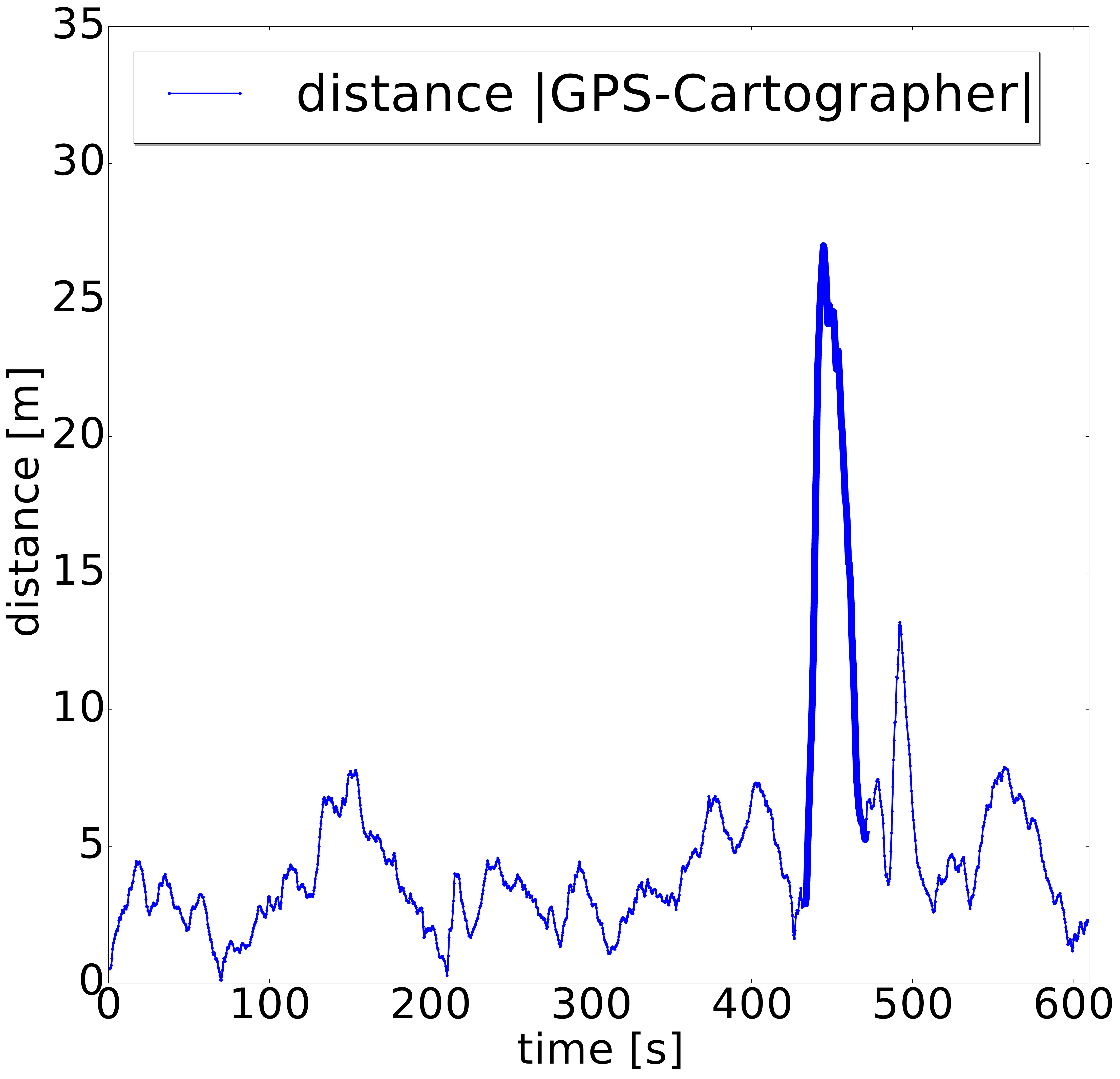}
  \end{subfigure}
  \caption{Comparison between GPS and Cartographer trajectories for Car Day 2 overlaid on top of satellite imagery. Note that the spike in error between Cartographer and GPS corresponds to the bolded section in the top right of the overlay on the left, and is largely due to GPS error. Best viewed in color.}
    \label{fig:driving_gps_vs_cartographer}
\end{figure*}

In order to provide a quantitative measure of the quality of the final pose, we align the positions with the GPS measurements, and provide both overlaid on top of satellite imagery for each outdoor sequence in the dataset, as well as the difference in position between the provided ground truth and GPS. Fig~\ref{fig:driving_gps_vs_cartographer} provides a sample overlay for Car Day 2, where the average error between Cartographer and the GPS is consistently around 5m without drift. This error is consistent amongst all of the outdoor driving sequences, where the overall average error is 4.7m, and is on a similar magnitude to the error expected from GPS. Note that the spike in error around 440 seconds is due to significant GPS error, and corresponds to the section in bold on the top right of the overlay.

In both cases, the extrinsic transform, represented as a $4\times 4$ homogenous transform matrix $^{\text{body}}\mathbf{H}_{\text{DAVIS}}$, for each sequence that takes a point from the left DAVIS frame to the body frame is then used to estimate the pose of the left DAVIS at time $t$ with respect to the first left DAVIS pose at time $t_0$:
\begin{align}
^{\text{DAVIS}(t_0)}\mathbf{H}_{\text{DAVIS}(t)}=&\notag\\{}^{\text{body}}\mathbf{H}_{\text{DAVIS}}^{-1}&{}^{\text{world}}\mathbf{H}_{\text{body}(t_0)}^{-1}{}^{\text{world}}\mathbf{H}_{\text{body}(t)}{}^{\text{body}}\mathbf{H}_{\text{DAVIS}}.
\label{eq:coordtransform}
\end{align}

\subsection{Depth Map Generation}
\begin{figure*}[h!]
\centering
  \includegraphics[trim=0 0 0 -1cm, width=0.6\linewidth]{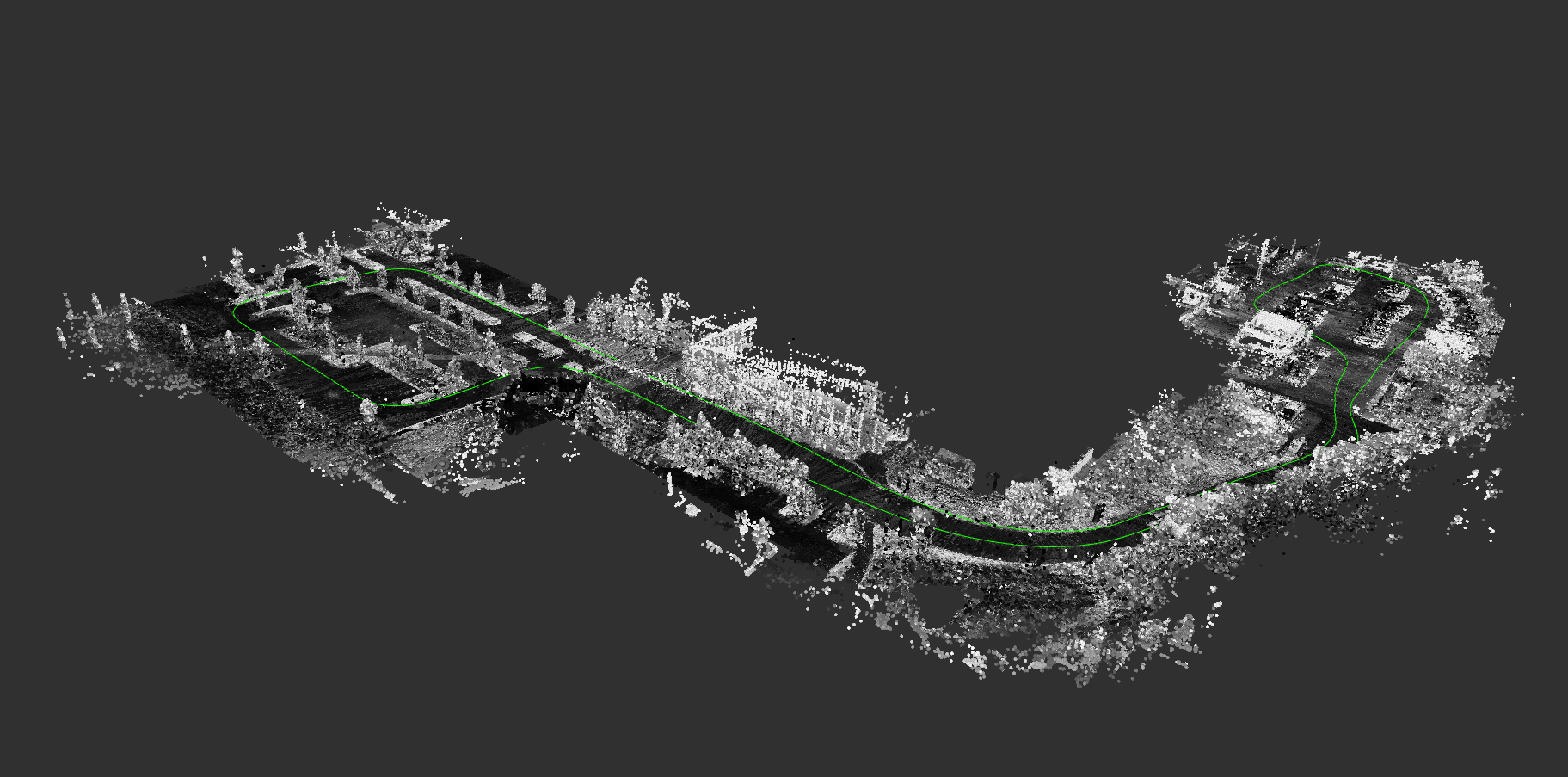}
  \includegraphics[trim=0 0 0 -1cm, width=0.325\linewidth]{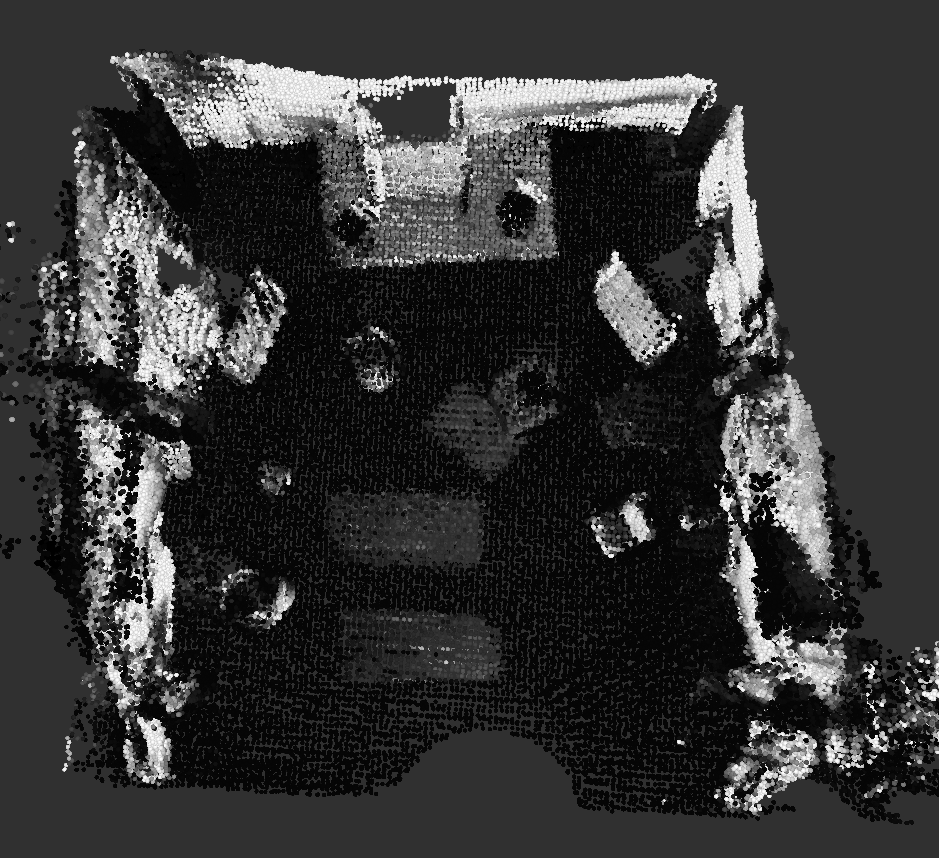}
  \caption{Sample maps generated for ground truth. \textbf{Left}: Full map from Car Day 1 sequence, trajectory in green. \textbf{Right}: Local map from the Hexacopter Indoor 3 sequence.}
  \label{fig:maps}
\end{figure*}

In each sequence where lidar is available, depth images for each DAVIS camera are generated for every lidar measurement. We first generate a local map by transforming each lidar pointcloud in a local window around the current measurement into the frame of the current measurement using the poses from LOAM. At each measurement, the window size is determined such that the distances between the current, and the first and last LOAM poses in the window are at least $d$ meters, and that there are at least $s$ seconds between the current, and first and last LOAM poses, where $d$ and $s$ are parameters tuned for each sequence. Examples of these maps can be found in Fig~\ref{fig:maps}. 

We then project each point, $\mathbf{p}$, in the resulting pointcloud into the image in each DAVIS camera, using the standard pinhole projection equation:
\begin{align}
\begin{pmatrix}
u & v & 1
\end{pmatrix}^T=&\mathbf{K}\Pi\left({}^{\text{body}}\mathbf{H}_{\text{DAVIS}} \begin{bmatrix}\mathbf{p} \\ 1\end{bmatrix}\right)
\intertext{where $\Pi$ is the projection function:}
\Pi\left(\begin{pmatrix}X & Y & Z & 1\end{pmatrix}^T\right)=& \begin{pmatrix}\frac{X}{Z} & \frac{Y}{Z} & 1\end{pmatrix}^T
\end{align}
and $\mathbf{K}$ is the camera intrinsics matrix for the rectified image (i.e. the top left $3\times 3$ of the projection matrix).

\begin{figure*}[h!]
\centering
  \includegraphics[width=0.45\linewidth]{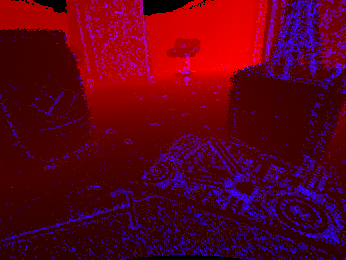}
  \includegraphics[width=0.45\linewidth]{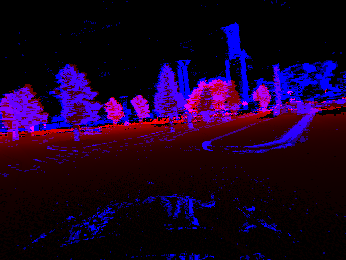}
  \caption{Depth images (red) with events overlaid (blue) from the Hexacopter Indoor 2 and Car Day 1 sequences. Note that parts of the image (black areas, particularly the top) have no depth due to the limited vertical field of view and range of the lidar. These parts are labeled as NaNs in the data. Best viewed in color.}
  \label{fig:depth_images}
\end{figure*}

Any points falling outside the image bounds are discarded, and the closest point at each pixel location in the image is used to generate the final depth map, examples of which can be found in Fig~\ref{fig:depth_images}. 

In addition, we also provide raw depth images without any undistortion by unrectifying and distorting the rectified depth images using the camera intrinsics and OpenCV.

\addtolength{\voffset}{10pt}
\section{CALIBRATION} \label{sec:calibration}
In this section, we describe the various steps performed to calibrate the intrinsic parameters of each DAVIS and VI-Sensor camera, as well as the extrinsic transformations between each of the cameras, IMUs and the lidar. All of the calibration results are provided in yaml form.

The camera intrinsics, stereo extrinsics, and camera-IMU extrinsics are calibrated using the Kalibr toolbox\footnote{\url{https://github.com/ethz-asl/kalibr}}\cite{kalibr1}, \cite{kalibr2}, \cite{kalibr3}, the extrinsics between the left DAVIS camera and Velodyne lidar are calibrated using the Camera and Range Calibration Toolbox\footnote{\url{http://www.cvlibs.net/software/calibration/}}\cite{lidarcamcalib}, and fine tuned manually, and the hand eye calibration between the mocap model pose in the motion capture world frame and the left DAVIS camera pose is performed using CamOdoCal\footnote{\url{https://github.com/hengli/camodocal}} \cite{camodocal}. To compensate for changes in the mounted rig, each calibration is repeated each day data was collected, and every time the sensing payload was modified. In addition to the calibration parameters, the raw calibration data for each day is also available on demand for users to perform their own calibration, if desired.
\subsection{Camera Intrinsic, Extrinsic and Temporal Calibration}
The camera intrinsics and extrinsics are estimated using a grid of AprilTags \cite{apriltags} that is moved in front of the sensor rig and calibrated using Kalibr. Each calibration provides the focal length and principal point of each camera as well as the distortion parameters and the extrinsics between the cameras. 

In addition, we calibrate the temporal offset between the DAVIS stereo pair and the VI Sensor by finding the temporal offset that maximizes the cross correlation between the magnitude of the gyroscope angular velocities from the IMUs of the left DAVIS and the VI Sensor. The timestamps for the VI Sensor messages in the dataset are then modified to compensate for this offset.
\subsection{Camera to IMU Extrinsic Calibration}
To calibrate the transformation between the camera and IMU, a sequence is recorded with the sensor rig moving in front of the AprilTag grid. The two calibration procedures are separated to optimize the quality of each individual calibration. The calibration sequences are once again run through Kalibr using the camera-IMU calibration to estimate the transformations between each camera and each IMU, given the prior intrinsic and camera-camera extrinsic calibrations.
\subsection{Motion Capture to Camera Extrinsic Calibration}
Each motion capture system provides the pose of the mocap model in the motion capture frame at 100Hz. However, the mocap model frame is not aligned with any camera frame, and so a further calibration is needed to obtain the pose of the cameras from the motion capture system.

The sensor rig was statically held in front of an Aprilgrid at a variety of different poses. At each pose at time $t_i$, the pose of the left DAVIS camera frame in the grid frame ${}^{\text{aprilgrid}}\mathbf{H}_{\text{DAVIS}(t_i)}$, as well as the pose of the mocap model (denoted body) in the mocap frame ${}^{\text{mocap}}\mathbf{H}_{\text{body}(t_i)}$, were measured. These poses were then used to solve the handeye calibration problem for the transform that transforms a point in the left DAVIS frame into the model frame ${}^{\text{body}}\mathbf{H}_{\text{DAVIS}}$:
\begin{align}
{}^{\text{body}(t_0)}\mathbf{H}_{\text{body}(t_i)}
{}^{\text{body}}\mathbf{H}_{\text{DAVIS}} =&{}^{\text{body}}\mathbf{H}_{\text{DAVIS}}
{}^{\text{DAVIS}(t_0)}\mathbf{H}_{\text{DAVIS}(t_i)}\notag\\
i =& 1,\ldots, n
\end{align}
where:
\begin{align}
{}^{\text{body}(t_0)}\mathbf{H}_{\text{body}(t_i)}=&{}^{\text{mocap}}\mathbf{H}_{\text{body}(t_0)}^{-1}{}^{\text{mocap}}\mathbf{H}_{\text{body}(t_i)}\\
{}^{\text{DAVIS}(t_0)}\mathbf{H}_{\text{DAVIS}(t_i)}=&{}^{\text{aprilgrid}}\mathbf{H}_{\text{DAVIS}(t_0)}^{-1}{}^{\text{aprilgrid}}\mathbf{H}_{\text{DAVIS}(t_i)}.
\end{align}
The optimization is performed using CamOdoCal, using the linear method in \cite{daniilidis1999hand}, and refining using a nonlinear optimization as described in \cite{camodocal}.
\subsection{Lidar to Camera Extrinsic Calibration}
\label{sec:lidartocamcalib}
The transformation that takes a point from the lidar frame to the left DAVIS frame was initially calibrated using the Camera and Range Calibration Toolbox \cite{lidarcamcalib}. Four large checkerboard patterns are placed to fill the field of view of the DAVIS cameras, and a single pair of images from each camera is recorded, along with a full lidar scan. 
The calibrator then estimates the translation and rotation that aligns the camera and lidar observations of the checkerboards.

However, we found that the reported transform had up to five pixels of error when viewing the projected depth images (Fig \ref{fig:depth_images}). In addition, as the lidar and cameras are not hardware time synchronized, there was occasionally a noticeable and constant time delay between the two sensors. To improve the calibration, we fixed the translation based on the values from the CAD models, and manually fine tuned the rotation and time offset in order to maximize the overlap between the depth and event images. For visual confirmation, we provide the depth images with events overlaid for each camera. The timestamps of the lidar messages provided in the dataset are compensated for the time offset.
\section{KNOWN ISSUES}
\subsection{Moving Objects}
The mapping used to generate the depth maps assumes that scenes are static, and typically does not filter out points on moving objects. As a result, the reported depth maps may have errors of up to two meters when tracking points on other cars, etc. However, these objects are typically quite rare compared to the total amount of data available. If desired, future work could involve classifying vehicles in the images and omitting these points from the depth maps.
\subsection{Clock Synchronization}
The motion capture and GPS are only synchronized to the rest of the system using the host computer's time. This may incur an offset between the reported timestamps and the actual measurement time. We record all measurements on one computer to minimize this effect. In addition, there may be some delay between a lidar point's measurement and the timestamp of the message due to the spin rate of the lidar.
\subsection{DVS Biasing}
Default biases for each camera were used when generating each sequence. However, it has been noted that, for the indoor flying sequences, the ratio of positive to negative events is higher than usual ($\sim$2.5-5x). At this point, we are unaware of what may have caused this imbalance, or whether tuning the biases would have balanced it. We note that the imbalance is particularly skewed over the speckled floor. We advise researchers using the polarities of the events to be aware of this imbalance when working with these sequences.
\addtolength{\textheight}{-50pt}
\section{CONCLUSION}
We present a novel dataset for stereo event cameras, on a number of different vehicles and in a number of different environments, with ground truth 6dof pose and depth images. We hope that this data can provide one standard on which new event based methods can be evaluated and compared.


\bibliographystyle{IEEEtran}
\bibliography{IEEEabrv,main,calibration,related_work}

\begin{thebibliography}{10}
\providecommand{\url}[1]{#1}
\csname url@rmstyle\endcsname
\providecommand{\newblock}{\relax}
\providecommand{\bibinfo}[2]{#2}
\providecommand\BIBentrySTDinterwordspacing{\spaceskip=0pt\relax}
\providecommand\BIBentryALTinterwordstretchfactor{4}
\providecommand\BIBentryALTinterwordspacing{\spaceskip=\fontdimen2\font plus
\BIBentryALTinterwordstretchfactor\fontdimen3\font minus
  \fontdimen4\font\relax}
\providecommand\BIBforeignlanguage[2]{{%
\expandafter\ifx\csname l@#1\endcsname\relax
\typeout{** WARNING: IEEEtran.bst: No hyphenation pattern has been}%
\typeout{** loaded for the language `#1'. Using the pattern for}%
\typeout{** the default language instead.}%
\else
\language=\csname l@#1\endcsname
\fi
#2}}

\bibitem{weikersdorfer2014event}
D.~Weikersdorfer, D.~B. Adrian, D.~Cremers, and J.~Conradt, ``Event-based 3{D
  SLAM} with a depth-augmented dynamic vision sensor,'' in \emph{Robotics and
  Automation (ICRA), 2014 IEEE International Conference on}.\hskip 1em plus
  0.5em minus 0.4em\relax IEEE, 2014, pp. 359--364.

\bibitem{rueckauer2016evaluation}
B.~Rueckauer and T.~Delbruck, ``Evaluation of event-based algorithms for
  optical flow with ground-truth from inertial measurement sensor,''
  \emph{Frontiers in neuroscience}, vol.~10, 2016.

\bibitem{barranco2016dataset}
F.~Barranco, C.~Fermuller, Y.~Aloimonos, and T.~Delbruck, ``A dataset for
  visual navigation with neuromorphic methods,'' \emph{Frontiers in
  neuroscience}, vol.~10, 2016.

\bibitem{eventcameradataset}
\BIBentryALTinterwordspacing
E.~Mueggler, H.~Rebecq, G.~Gallego, T.~Delbruck, and D.~Scaramuzza, ``The
  event-camera dataset and simulator: Event-based data for pose estimation,
  visual odometry, and {SLAM},'' \emph{The International Journal of Robotics
  Research}, vol.~36, no.~2, pp. 142--149, 2017. [Online]. Available:
  \url{http://dx.doi.org/10.1177/0278364917691115}
\BIBentrySTDinterwordspacing

\bibitem{binas2017ddd17}
J.~Binas, D.~Niel, S.-C. Liu, and T.~Delbruck, ``Ddd17: End-to-end davis
  driving dataset,'' 2017.

\bibitem{kogler2010address}
J.~Kogler, C.~Sulzbachner, F.~Eibensteiner, and M.~Humenberger, ``Address-event
  matching for a silicon retina based stereo vision system,'' in \emph{4th Int.
  Conference from Scientific Computing to Computational Engineering}, 2010, pp.
  17--24.

\bibitem{kogler2011event}
J.~Kogler, M.~Humenberger, and C.~Sulzbachner, ``Event-based stereo matching
  approaches for frameless address event stereo data,'' \emph{Advances in
  Visual Computing}, pp. 674--685, 2011.

\bibitem{piatkowska2014cooperative}
E.~Piatkowska, A.~N. Belbachir, and M.~Gelautz, ``Cooperative and asynchronous
  stereo vision for dynamic vision sensors,'' \emph{Measurement Science and
  Technology}, vol.~25, no.~5, p. 055108, 2014.

\bibitem{Firouzi2016}
\BIBentryALTinterwordspacing
M.~Firouzi and J.~Conradt, ``Asynchronous event-based cooperative stereo
  matching using neuromorphic silicon retinas,'' \emph{Neural Processing
  Letters}, vol.~43, no.~2, pp. 311--326, Apr 2016. [Online]. Available:
  \url{https://doi.org/10.1007/s11063-015-9434-5}
\BIBentrySTDinterwordspacing

\bibitem{piatkowska2017improved}
E.~Piatkowska, J.~Kogler, N.~Belbachir, and M.~Gelautz, ``Improved cooperative
  stereo matching for dynamic vision sensors with ground truth evaluation,'' in
  \emph{Computer Vision and Pattern Recognition Workshops (CVPRW), 2017 IEEE
  Conference on}.\hskip 1em plus 0.5em minus 0.4em\relax IEEE, 2017, pp.
  370--377.

\bibitem{rogister2012asynchronous}
P.~Rogister, R.~Benosman, S.-H. Ieng, P.~Lichtsteiner, and T.~Delbruck,
  ``Asynchronous event-based binocular stereo matching,'' \emph{IEEE
  Transactions on Neural Networks and Learning Systems}, vol.~23, no.~2, pp.
  347--353, 2012.

\bibitem{carneiro2013event}
J.~Carneiro, S.-H. Ieng, C.~Posch, and R.~Benosman, ``Event-based 3{D}
  reconstruction from neuromorphic retinas,'' \emph{Neural Networks}, vol.~45,
  pp. 27--38, 2013.

\bibitem{camunas2014use}
L.~A. Camu{\~n}as-Mesa, T.~Serrano-Gotarredona, S.~H. Ieng, R.~B. Benosman, and
  B.~Linares-Barranco, ``On the use of orientation filters for 3{D}
  reconstruction in event-driven stereo vision,'' \emph{Frontiers in
  neuroscience}, vol.~8, 2014.

\bibitem{benosman2011asynchronous}
R.~Benosman, S.-H. Ieng, P.~Rogister, and C.~Posch, ``Asynchronous event-based
  {H}ebbian epipolar geometry,'' \emph{IEEE Transactions on Neural Networks},
  vol.~22, no.~11, pp. 1723--1734, 2011.

\bibitem{zou2016context}
D.~Zou, P.~Guo, Q.~Wang, X.~Wang, G.~Shao, F.~Shi, J.~Li, and P.-K. Park,
  ``Context-aware event-driven stereo matching,'' in \emph{Image Processing
  (ICIP), 2016 IEEE International Conference on}.\hskip 1em plus 0.5em minus
  0.4em\relax IEEE, 2016, pp. 1076--1080.

\bibitem{schraml2015event}
S.~Schraml, A.~Nabil~Belbachir, and H.~Bischof, ``Event-driven stereo matching
  for real-time 3{D} panoramic vision,'' in \emph{Proceedings of the IEEE
  Conference on Computer Vision and Pattern Recognition}, 2015, pp. 466--474.

\bibitem{zhu2017event}
A.~Z. Zhu, N.~Atanasov, and K.~Daniilidis, ``Event-based feature tracking with
  probabilistic data association,'' in \emph{Robotics and Automation (ICRA),
  2017 IEEE International Conference on}.\hskip 1em plus 0.5em minus
  0.4em\relax IEEE, 2017, pp. 4465--4470.

\bibitem{tedaldi2016feature}
D.~Tedaldi, G.~Gallego, E.~Mueggler, and D.~Scaramuzza, ``Feature detection and
  tracking with the dynamic and active-pixel vision sensor ({DAVIS}),'' in
  \emph{Event-based Control, Communication, and Signal Processing (EBCCSP),
  2016 Second International Conference on}.\hskip 1em plus 0.5em minus
  0.4em\relax IEEE, 2016, pp. 1--7.

\bibitem{kueng2016low}
B.~Kueng, E.~Mueggler, G.~Gallego, and D.~Scaramuzza, ``Low-latency visual
  odometry using event-based feature tracks,'' in \emph{Intelligent Robots and
  Systems (IROS), 2016 IEEE/RSJ International Conference on}.\hskip 1em plus
  0.5em minus 0.4em\relax IEEE, 2016, pp. 16--23.

\bibitem{zhuevent}
A.~Z. Zhu, N.~Atanasov, and K.~Daniilidis, ``Event-based visual inertial
  odometry,'' in \emph{Proceedings of the IEEE Conference on Computer Vision
  and Pattern Recognition}, 2017, pp. 5391--5399.

\bibitem{gallego2017accurate}
G.~Gallego and D.~Scaramuzza, ``Accurate angular velocity estimation with an
  event camera,'' \emph{IEEE Robotics and Automation Letters}, vol.~2, no.~2,
  pp. 632--639, 2017.

\bibitem{kim2016real}
H.~Kim, S.~Leutenegger, and A.~J. Davison, ``Real-time 3{D} reconstruction and
  6-dof tracking with an event camera,'' in \emph{European Conference on
  Computer Vision}.\hskip 1em plus 0.5em minus 0.4em\relax Springer, 2016, pp.
  349--364.

\bibitem{rebecq2017evo}
H.~Rebecq, T.~Horstschaefer, G.~Gallego, and D.~Scaramuzza, ``{EVO}: A
  geometric approach to event-based 6-dof parallel tracking and mapping in real
  time,'' \emph{IEEE Robotics and Automation Letters}, vol.~2, no.~2, pp.
  593--600, 2017.

\bibitem{rebecq2017real}
H.~Rebecq, T.~Horstschaefer, and D.~Scaramuzza, ``Real-time visual-inertial
  odometry for event cameras using keyframe-based nonlinear optimization,'' in
  \emph{British Machine Vis. Conf.(BMVC)}, vol.~3, 2017.

\bibitem{mueggler2017continuous}
E.~Mueggler, G.~Gallego, H.~Rebecq, and D.~Scaramuzza, ``Continuous-time
  visual-inertial trajectory estimation with event cameras,'' \emph{arXiv
  preprint arXiv:1702.07389}, 2017.

\bibitem{brandli2014240}
C.~Brandli, R.~Berner, M.~Yang, S.-C. Liu, and T.~Delbruck, ``A 240$\times$ 180
  130 d{B} 3 $\mu$s latency global shutter spatiotemporal vision sensor,''
  \emph{IEEE Journal of Solid-State Circuits}, vol.~49, no.~10, pp. 2333--2341,
  2014.

\bibitem{visensor}
J.~Nikolic, J.~Rehder, M.~Burri, P.~Gohl, S.~Leutenegger, P.~T. Furgale, and
  R.~Siegwart, ``A synchronized visual-inertial sensor system with {FPGA}
  pre-processing for accurate real-time {SLAM},'' in \emph{2014 IEEE
  International Conference on Robotics and Automation (ICRA)}, May 2014, pp.
  431--437.

\bibitem{cartographer}
W.~Hess, D.~Kohler, H.~Rapp, and D.~Andor, ``Real-time loop closure in 2{D
  LIDAR SLAM},'' in \emph{Robotics and Automation (ICRA), 2016 IEEE
  International Conference on}.\hskip 1em plus 0.5em minus 0.4em\relax IEEE,
  2016, pp. 1271--1278.

\bibitem{loam}
J.~Zhang and S.~Singh, ``{LOAM}: Lidar odometry and mapping in real-time.'' in
  \emph{Robotics: Science and Systems}, vol.~2, 2014.

\bibitem{kalibr1}
P.~Furgale, J.~Rehder, and R.~Siegwart, ``Unified temporal and spatial
  calibration for multi-sensor systems,'' in \emph{Intelligent Robots and
  Systems (IROS), 2013 IEEE/RSJ International Conference on}.\hskip 1em plus
  0.5em minus 0.4em\relax IEEE, 2013, pp. 1280--1286.

\bibitem{kalibr2}
P.~Furgale, T.~D. Barfoot, and G.~Sibley, ``Continuous-time batch estimation
  using temporal basis functions,'' in \emph{Robotics and Automation (ICRA),
  2012 IEEE International Conference on}.\hskip 1em plus 0.5em minus
  0.4em\relax IEEE, 2012, pp. 2088--2095.

\bibitem{kalibr3}
J.~Maye, P.~Furgale, and R.~Siegwart, ``Self-supervised calibration for robotic
  systems,'' in \emph{Intelligent Vehicles Symposium (IV), 2013 IEEE}.\hskip
  1em plus 0.5em minus 0.4em\relax IEEE, 2013, pp. 473--480.

\bibitem{lidarcamcalib}
A.~Geiger, F.~Moosmann, Ömer Car, and B.~Schuster, ``Automatic camera and
  range sensor calibration using a single shot,'' in \emph{International
  Conference on Robotics and Automation (ICRA)}, St. Paul, USA, May 2012.

\bibitem{camodocal}
L.~Heng, B.~Li, and M.~Pollefeys, ``Camodocal: Automatic intrinsic and
  extrinsic calibration of a rig with multiple generic cameras and odometry,''
  in \emph{Intelligent Robots and Systems (IROS), 2013 IEEE/RSJ International
  Conference on}.\hskip 1em plus 0.5em minus 0.4em\relax IEEE, 2013, pp.
  1793--1800.

\bibitem{apriltags}
E.~Olson, ``Apriltag: A robust and flexible visual fiducial system,'' in
  \emph{Robotics and Automation (ICRA), 2011 IEEE International Conference
  on}.\hskip 1em plus 0.5em minus 0.4em\relax IEEE, 2011, pp. 3400--3407.

\bibitem{daniilidis1999hand}
K.~Daniilidis, ``Hand-eye calibration using dual quaternions,'' \emph{The
  International Journal of Robotics Research}, vol.~18, no.~3, pp. 286--298,
  1999.

\end{thebibliography}

\end{document}